\documentclass[letterpaper, 10 pt, conference]{ieeeconf}  
\usepackage{times}
\usepackage{epsfig}
\usepackage{graphicx}
\usepackage{mathrsfs, amsmath}
\usepackage{amssymb}
\usepackage{hyperref}
\usepackage{booktabs}
\usepackage[font=small,labelfont=bf]{caption}
\usepackage{url}
\usepackage[table,dvipsnames]{xcolor}
\usepackage{verbatim}
\usepackage{listings}
\usepackage{multirow}

\graphicspath{{./fig/}}
\definecolor{mygray}{gray}{0.8}

\IEEEoverridecommandlockouts                              

\title{\LARGE \bf Benchmarking of Deep Reinforcement Learning Algorithms for Vision-based Robotics}

\author{Swagat Kumar, Hayden Sampson and Ardhendu Behera
  \thanks{The authors are with the department of Computer Science, Edge Hill University,
    Ormskirk, UK L39 4QP. email: {\tt\small \{kumars,haydens,beheraa\}@edgehill.ac.uk}.} 
}

\begin{document}

\maketitle
\thispagestyle{empty}
\pagestyle{empty}

\begin{abstract}
  This paper presents a benchmarking study of some of the
  state-of-the-art reinforcement learning algorithms used for solving
  two simulated vision-based robotics problems.  The algorithms
  considered in this study include  soft actor-critic (SAC), proximal
  policy optimization (PPO), interpolated policy gradients (IPG), and
  their variants with Hindsight Experience replay (HER). The
  performances of these algorithms are compared against PyBullet's two
  simulation environments known as  \texttt{KukaDiverseObjectEnv} and
  \texttt{RacecarZEDGymEnv} respectively. The state observations in
  these environments are available in the form of RGB images and the
  action space is continuous, making them difficult to solve. A number
  of strategies are suggested to provide intermediate hindsight goals
  required for implementing HER algorithm on these problems which are
  essentially single-goal environments. In addition, a number of
  feature extraction architectures are proposed to incorporate spatial
  and temporal attention in the learning process. Through rigorous
  simulation experiments, the improvement achieved with these
  components are established. To the best of our knowledge, such a
  benchmarking study is not available for the above two vision-based
  robotics problems making it a novel contribution in the field.    
\end{abstract} 

\section{Introduction}\label{sec:intro}
Vision-based Robots use visual feedback to guide their motion. The
underlying problem, also known as hand-eye coordination 
\cite{hager1995robot} or visual servoing \cite{chaumette2016visual},
is considered to be difficult due to factors such as
non-linearity and uncertainty of camera and robot models, sensor
noise, mathematical complexity of extracting geometrical features from
images and difficulty in deriving closed-form analytical control
equations in terms of these image features etc. Traditional model-based
methods addressed some of these concerns by adopting soft computing
approaches \cite{suh1994fuzzy} \cite{kumar2010visual}
\cite{fukuda1998soft}. The recent success of deep learning methods in
computer vision has greatly enhanced the capabilities of these
model-based methods \cite{bateux2018training}
\cite{saxena2017exploring} \cite{lee2017learning}
\cite{li2020learning}. In contrast, reinforcement learning (RL) offers a
model-free alternative to solve this problem directly from
input-output data, thereby overcoming many of the above problems. It
has the potential to create truly autonomous agents that can learn
skills on its own without needing human intervention. An RL agent learns the
desired behaviour, over time, through trial and error while repeatedly
interacting with its environment \cite{sutton2018reinforcement}. The
agent achieves this by taking actions to maximize the \emph{cumulative
discounted future reward} for a given task 
while balancing \emph{exploration} (of new possibilities) and
\emph{exploitation} (of past experiences). This cumulative discounted
reward function, represented as Q or value function, is used to
evaluate a given action, and is not known a priori.  Depending on how
this function is estimated and desirable actions are derived from it,
the RL-based methods can be broadly classified into two categories:
\emph{value-based} methods and \emph{policy-based} methods.
Value-based methods aim at estimating the Q-function and then derive
action from this by using a greedy policy. On the other hand,
policy-based methods directly estimate the policy function by
maximizing a given objective function. The traditional Q-learning
algorithm  estimates the Q function iteratively by using an
approximate dynamic programming formulation based on Bellman's
equation starting from an initial estimate
\cite{barto1995reinforcement}. The original Q-learning algorithm could
be applied to problems with discrete state (observation) and action
spaces, and suffer from the \emph{curse-of-dimensionality} problem
with higher dimensions and range of values. This limitation can be
overcome by using a deep network to estimate Q function from arbitrary
observation inputs, thereby, greatly enhancing the capabilities of RL
algorithms. The resulting approach is known as Deep Q Networks (DQN)
\cite{van2016deep} \cite{sewak2019deep} which has been applied
successfully to a wide range of problems while achieving superhuman
level performances in a few cases, such as ATARI video games
\cite{mnih2013playing}, Go \cite{holcomb2018overview} etc.  The
success of DQN has spawned a new research field known as deep
reinforcement learning (DRL) attracting a large following of
researchers. Readers are referred to \cite{arulkumaran2017deep} for a
survey of this field. The DQN models were subsequently extended to
continuous action spaces by using policy gradient methods that used a
parameterized policy function to maximize DQN output using gradient
ascent methods \cite{lillicrap2015continuous}
\cite{duan2016benchmarking}. This has opened the doors for solving
various robotics problems that use continuous values such as joint
angle, joint velocities or motor torques as input. Since then, a
number of methods have been proposed to improve the performance of RL
algorithms and have been applied successfully to different robotic
problems - manipulation \cite{gu2017deep} \cite{nguyen2019review},
grasping \cite{quillen2018deep} \cite{joshi2020robotic}, navigation
\cite{yue2019experimental} etc.  

Since RL algorithms vary greatly in terms of model complexity, choice
of hyper-parameters and architectures, it becomes necessary to
benchmark them against standard problems to assess their performances.
One such effort was made for a robotic grasp problem in
\cite{quillen2018deep} where authors compared the performance of three
off-policy methods including DDPG \cite{lillicrap2015continuous} on a
simulated Kuka environment created using PyBullet \cite{pybullet}. A
similar benchmarking was done for continuous control problems in
\cite{duan2016benchmarking} where authors compared the performance of
several off-policy and on-policy methods on several simulated
environments created using MuJoCo \cite{mujoco}. A benchmarking of
reinforcement learning algorithms on real-world robots was carried out
in \cite{mahmood2018benchmarking} that included algorithms such as
TRPO \cite{schulman2015trust} and PPO \cite{schulman2017proximal}. 

The focus of this paper is to benchmark some of the newer deep
reinforcement learning algorithms such as Soft Actor Critic (SAC)
\cite{haarnoja2018soft}, Proximal Policy Optimization (PPO)
\cite{schulman2017proximal} and Interpolated Policy Gradients (IPG)
\cite{gu2017interpolated} for solving vision-based robotic
problems in an end-to-end fashion. The problems used for this study
include two PyBullet \cite{pybullet} simulation environment, namely,
\texttt{KukaDiverseObject} and \texttt{RacecarZEDGym}. Both
environments provide observations in the form of RGB images and take
continuous action inputs which makes these two problems more difficult
compared to the versions where observations are available as floating
point vectors. We then demonstrate that the performance of these
algorithms can be further improved by using Hindsight experience
replay (HER) \cite{andrychowicz2017hindsight} where the problem of
sparse rewards is partially addressed by providing intermediate goals.
HER is shown to be more effective for multi-goal problems such as
MuJoCo's Fetch environment \cite{plappert2018multi} or PyBullet's
Panda robot \cite{gallouedec2021multi}. Both of these two environments
do not provide image observations making them unsuitable for this
study. A number of strategies are proposed to provide intermediate
goals for the above two environments to facilitate implementation of
HER algorithm and show improvement in performance achieved.  Finally,
a number of architectures are presented to incorporate spatial and
temporal attention into the learning process. Attention mechanisms
\cite{vaswani2017attention} have been shown to improve the inference
capability of a model by focussing on relevant parts and ignoring the
irrelevant parts of the input observation. While the application of
attention in reinforcement learning is not new, it has not been
applied to the context of vision-based robotics problems considered in
this paper. The performance improvement achieved by 
incorporating these concepts are demonstrated through rigorous
simulation experiments which will be discussed later in this paper. 

In short, the novel contributions made in this paper are as follows:
\begin{itemize}
  \item A benchmarking of three RL algorithms, namely, SAC, PPO, IPG
    and their corresponding HER-variants have been carried out for two
    simulated vision-based robotics problem. This is the first-time
    such an effort has been made for these two simulated PyBullet
    environments.

  \item Several strategies are proposed to implement HER algorithm on these two environments which are
    essentially single-goal problems and do not provide intermediate
    goals. These ideas could be easily applied to other single-goal
    environments thereby extending the applicability of HER
    algorithms. 

  \item Several architectures are proposed for incorporating spatial
    and temporal attention during the feature extraction process
    leading to improved learning performance. Through experiments, it
    is established that stacked frames along with attention and LSTM
    layers can provide superior learning performance. The application
    of attention in the present context of vision-based robotics is
    new and has not been reported so far. 
\end{itemize}

The rest of this paper is organized as follows. The details of RL model
architectures, algorithms and methods are discussed in Section
\ref{sec:meth}. The details of experiments and analysis of results are
presented in Section \ref{sec:expt}. The conclusion and direction of
future research is provided in Section \ref{sec:conc}.

\section{Problem Environments}\label{sec:prob}

\begin{figure*}[!t]
  \resizebox{0.5\textwidth}{!}{
  \begin{tabular}{cc}
    \includegraphics[scale=0.15]{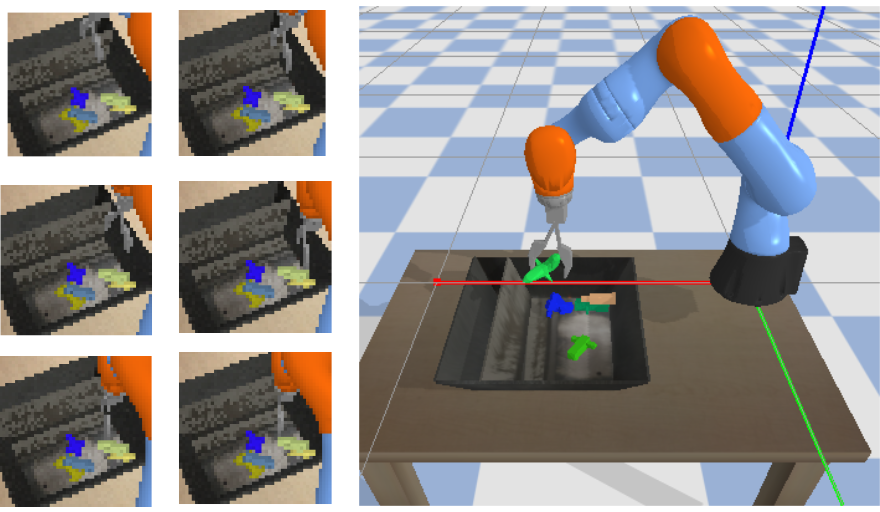} &
    \includegraphics[scale=0.15]{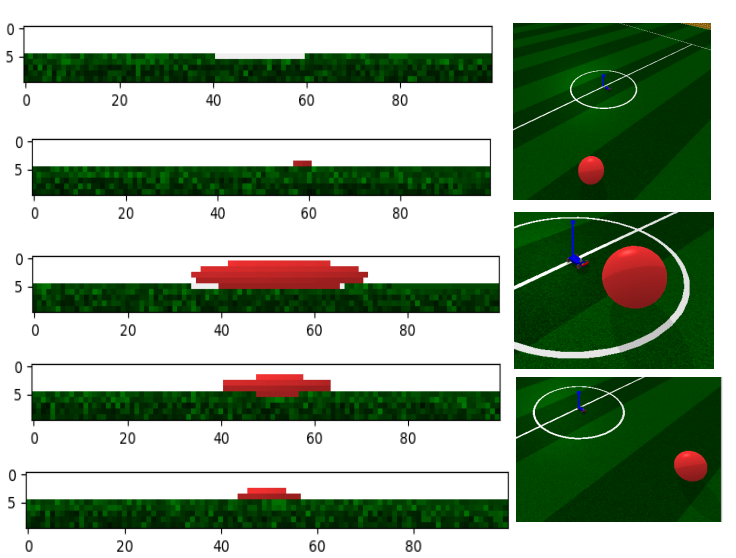} \\
    {\scriptsize (a) \tt{KukaDiverseObjectEnv}} & {\scriptsize (b) \tt{RacecarZEDGymEnv}}
  \end{tabular}}
  \hfill
  \resizebox{0.5\textwidth}{!}{
    \begin{tabular}{|p{3cm}|c|c|c|} \hline
     Parameters & {\small \tt{KukaDiverseObjectEnv}} & {\small \tt{RacecarZEDGymENV}}  \\ \hline
    Observation shape $s$ & (48, 48, 3) &  (10, 100, 4) \\ \hline
    Action $a$ & (3,) & (2,)  \\ \hline
    Action Range & [-1, 1] & [-1, 1] \\ \hline
    Reward Range $r$ & (-inf, inf)  & (-inf, inf)  \\ \hline
    Max steps / episode & 20 & 100 \\ \hline
    \multicolumn{3}{c}{} \\
    \multicolumn{3}{c}{ (c) Environment parameters}
  \end{tabular}}
  \caption{\footnotesize PyBullet Simulation Environment used for benchmarking RL
  algorithms. The input observation is available in the form of RGB
images and the action input are continuous floating point values. The
images for the racecar environment are resized to (40,40,3) before
use. }
  \label{fig:sim_env}
\end{figure*}

The objective of this work is to benchmark a few deep reinforcement
learning algorithms, namely, soft actor critic (SAC)
\cite{haarnoja2018soft}, proximal policy optimization (PPO)
\cite{schulman2017proximal}, interpolated policy gradients (IPG)
\cite{gu2017interpolated} and hindsight experience replay (HER)
\cite{andrychowicz2017hindsight} on two vision-based robotics problem.
The data intensive nature of reinforcement learning necessitates use
of simulated environments for training models. Two PyBullet
\cite{pybullet} environments namely \texttt{KukaDiverseObjectEnv} and
\texttt{RacecarZEDGymEnv} are used for implementing different RL
algorithms. In the first environment, a Kuka robotic arm is used to
grasp different kinds of objects from a bin. The environment gives a
reward of 1 when one of the objects is picked up beyond a pre-defined
height. The maximum reward for a given episode is 1 for the kuka
environment. The second environment consists of a robotic car in a
stadium and it is expected to reach a given target ball location by
generating linear and angular velocities. The agent receives a
floating point reward for each step in the episode. The input
observation for both environments is available in the form of RGB
images and the system accepts continuous floating point values as
actions. A snapshot of the environment rendering and corresponding
variables are shown in Figure \ref{fig:sim_env}. These two
environments are selected as they are freely available and fall within
the scope of vision-based robotics which is the focus of this work.
Our algorithms work equally well with other environments such as
\texttt{FetchReach} from  Mujoco \cite{mujoco}. However, the results
have not been included in this study because of two reasons - first,
they are proprietary and require purchasing license for long-term use
and secondly, image observations are not available for this
environment.

\section{Method}\label{sec:meth}
This section provides the details of models, architectures and
algorithms used for this benchmarking study as described in the
following subsections.

\subsection{The RL formulation}
An RL agent in state $s_t=s$ at time $t$ takes an action $a_t=a$
according to its policy $\pi(a|s)$. The state transitions to state
$s_{t+1}=s'$ at $t+1$ while receiving a reward $r(s_t,a_t) = r$ from
the environment. The terminal state of an episode is represented by a
boolean variable $d$. The goal of the reinforcement learning is to
learn a policy that will maximize the discounted cumulative future
returns given by $J(\theta)= \mathop{\mathbb{E}}_{s,a\sim
\pi}[\sum_{t=0}^\infty\gamma^t r(s_t,a_t)]$ where $\gamma$ is the
discount factor. This RL problem is solved in this paper using an
actor-critic model \cite{grondman2012survey} \cite{konda2000actor}
that uses two different deep networks - an actor network for learning
the policy function $\pi_\theta(a|s)$ and a critic network for
estimating the value function $V_w(s)$ or the Q-function $Q_w(s,a)$
where $\theta$ and $w$ are the trainable parameters. The critic is
trained to minimize the time-delay (TD) error function (also known as
mean square Bellman error (MBSE) given by:
\begin{equation}
  L(w_t) = \mathop{\mathbb{E}} [y - Q(s,a;w_t)]²
  \label{eq:critic_loss}
\end{equation}
where the target signal $y$ is obtained from the recursive Bellman's equation given by: 
\begin{equation}
  y = \mathop{\mathbb{E}}_{s'\in S} [r + \gamma \max_{a'\in \pi(s')} Q(s', a'; w_{t-1})]
  \label{eq:critic_target}
\end{equation}
The training process for critic network is similar to that of DQN
\cite{mnih2013playing} that uses off-policy experience replay to
improve the sample efficiency. The experience tuples $(s,a,r,s',d)$
are stored in a replay buffer $\mathscr{B}$ and then sampled in
mini-batches during the training process. The actor network, on the
other hand, is trained using a policy gradient method where the model
parameters $\theta$ are updated so as to maximize the critic output.
For instance, DDPG \cite{lillicrap2015continuous} optimize a
continuous deterministic policy $\pi_\theta(a_t|s_t) = \delta(a_t =
\mu_\theta(s_t))$ by applying the gradient ascent to the critic output
directly as shown below:
\begin{equation}
  \theta \leftarrow \theta + \alpha \mathop{\mathbb{E}}_{s,\pi \sim \mathscr{B}} [\nabla_\theta Q_w(s,\mu_\theta(s))]
  \label{eq:dpg}
\end{equation}
where $\alpha$ is the learning rate. DDPG uses target networks to stabilize the
learning process which are updated at a slower rate compared to
original models by using Polyak averaging
\cite{polyak1992acceleration}.  DDPG is known to be sensitive to the
choices of parameters and sometimes overestimates Q-values. This is
remedied in Twin-delayed DDPG (TD3)  algorithm \cite{dankwa2019twin}
that uses clipped double-Q learning, delayed policy updates and target
policy smoothing to improve the learning performance.  Many other
on-policy and off-policy methods have been reported in literature to
provide superior learning performances. Some of these methods are
briefly discussed in the following subsections.

\subsection{Model Architecture} \label{sec:model_arch}
\begin{figure}[!t]
  \centering
    \includegraphics[scale=0.25]{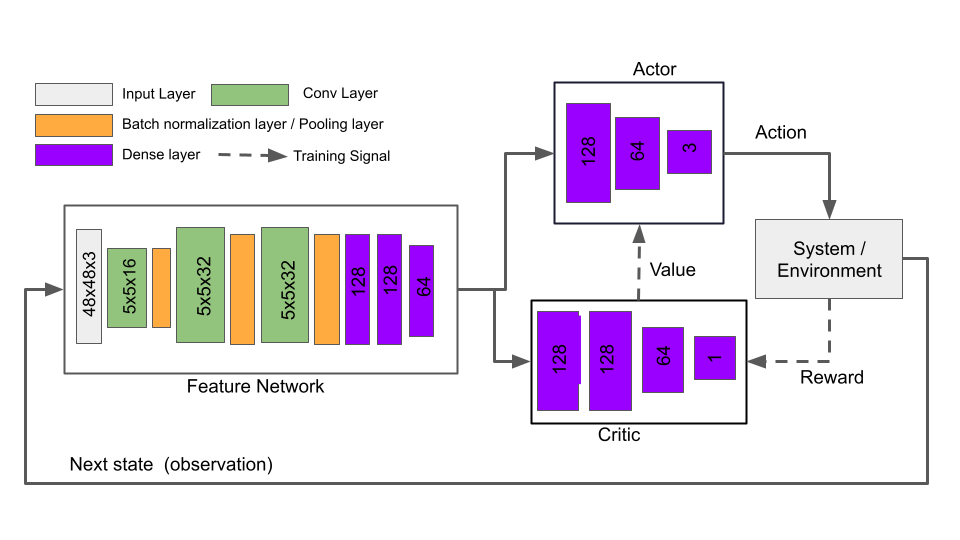}  
  \caption{Actor-Critic Architecture used for implementing RL
  algorithms. The feature network is shared by both actor and critic
models. Actor network approximates the policy function required for
producing actions required for transitioning to next states. The state
observations are made available in the form of RGB images. Critic
estimates value function $V(s)$ or Q-function  $Q(s,a)$.}
  \label{fig:net_arch}
\end{figure}

An overview of actor-critic model used in this paper is shown
in Figure \ref{fig:net_arch}. It consists of an actor network and a
critic network. The actor network approximates the policy function
required for producing actions for a given state observation. The
critic network, on the other hand, evaluates the actor network by
estimating the value function $V(s)$ or the Q function $Q(s,a)$. Both
of these networks share a common feature network which is used for
extracting features from the input RGB images. The feature network is
essentially a convolutional neural network (CNN). The number and size
of convolutional and dense layers may vary from problem to problem.
The feature network may include attention layers, and LSTM layers if
required as will be discussed later in this paper. It may also be
configured to receive a single image or a stack of frames as input.
The RL algorithms used for solving the above problem is described next
in this section. 

\subsection{Soft Actor-Critic Algorithm} \label{sec:sac}
Soft Actor Critic (SAC) \cite{haarnoja2018soft} algorithm optimizes a
stochastic policy in an off-policy manner. It uses a concept called
entropy regularization where the policy is trained to maximize a
trade-off between expected return and entropy in the policy.  In
addition, it borrows clipped double-Q trick from TD3 for stabilizing
the training process.  The target signal needed for critic update is
given by \begin{equation*} y = r + \gamma (1-d)
  \left(\min_{i=1,2}Q_{targ,i}(s',a') - \alpha \log
  \pi_\theta(a'|s')\right) \label{eq:sac_target} \end{equation*} where
the action values $a'$ are sampled from the current policy
$\pi_\theta(.|s)$.  Hence, the critic is trained by using
gradient-descent to minimize the MSBE equation \eqref{eq:critic_loss}
over mini-batches sampled from the replay buffer.  The policy
parameters are updated by applying gradient-ascent to maximize the
entropy-regularized value function.  The corresponding gradient term
may be written as follows:
\begin{equation*}
  \nabla_\theta\frac{1}{|\mathscr{B}|}\sum_{s\in B} \left(\min_{i=1,2}Q_{w_i}(s,\tilde{a}_\theta(s)) - \alpha \log \pi_\theta(\tilde{a}_\theta(s)|s)\right)
  \label{eq:sac_policy_update}
\end{equation*}
where $\tilde{a}_\theta(s)$ is a sample from $\pi_\theta(.|s)$ which is differentiable with respect to $\theta$.

\subsection{Proximal Policy Optimization} \label{sec:ppo}
On-policy methods use the likelihood ratio policy gradient
given below for policy optimization: \begin{equation} \nabla_\theta
  J(\theta) = \mathop{\mathbb{E}}[\nabla_\theta \log
  \pi_\theta(a|s)\hat{A}(s,a)] \label{eq:lrpgrad} \end{equation} where
$\hat{A}(s,a)$ is the estimate of advantage function. Proximal Policy Optimization (PPO) \cite{schulman2017proximal}, like
its predecessor TRPO \cite{schulman2015trust}, tries to maximize
the policy improvement step while keeping the new policy close to the
old policy.  It simplifies the TRPO by using a clipped version of
objective function which is much easier to implement. The policy
update equation is given by:
\begin{equation}
  \label{eq:ppo_policy_update}
  \begin{split}
    &\theta_{k+1} = \arg \max_\theta \frac{1}{|\mathscr{D}_k|T} \sum_{\tau \in \mathscr{D}_k} \sum_{t=0}^T \\
      & \quad \min  \left(\frac{\pi_\theta(a_t|s_t)}{\pi_{\theta_k}(a_t|s_t)} A^{\pi_{\theta_k}}(s_t, a_t), 
        g(\epsilon, A^{\pi_{\theta_k}}(s_t,a_t))\right) 
  \end{split}
\end{equation}
where
\[g(\epsilon, A) = \left\{
               \begin{array}{l}
                  (1+\epsilon)A, A>0 \\ 
                  (1-\epsilon)A, A<0 
               \end{array} 
               \right. \]
is the clipped advantage function used to constrain the new policy.
$\hat{A}=A^{\pi_\theta}$ is the generalized advantage estimate (GAE)
\cite{schulman2015high} obtained from the current trajectories stored
in an on-policy buffer $\mathscr{D}_k$.  The critic uses a gradient
descent algorithm to minimize the mean square Bellman error (MSBE) as
described before.  Compared to off-policy methods, on-policy gradient
methods tend to be stable and relatively easier to implement. However,
on-policy methods are known to be highly data inefficient as they look
into the data only once. 

\subsection{Interpolated Policy Gradient} \label{sec:ipg}
Interpolated policy gradient (IPG) \cite{gu2017interpolated} combines
the benefits of both off-policy and on-policy algorithms to provide
superior learning performance. Specifically, it mixes likelihood ratio
gradient $\hat{Q}$ and deterministic gradient through an off-policy
fitted critic $Q_w$. It uses the parameter $\nu$ to trade-off bias and
variance directly and a control variate to further reduce the
estimator variance. The overall policy gradient can be written as: 
\begin{equation}
  \begin{split}
  \nabla_\theta J(\theta) & \approx (1-\nu) \mathop{\mathbb{E}}_{\rho^\pi, \pi}[\nabla_\theta \log \pi_\theta (a_t|s_t)  (\hat{A}(s_t,a_t)  -  \\ 
& \quad \quad  A_w^\pi(s_t,a_t))] + \mathop{\mathbb{E}}_{\rho^\beta}[\nabla_\theta \bar{Q}_w^\pi(s_t)]
\end{split}
  \label{eq:ipg}
\end{equation}
where $\rho^\beta$ refers to off-policy state sampling and $\rho^\pi$
refers to on-policy state sampling. $A_w^\pi(s_t, a_t)$ is the
advantage estimated with a baseline value function estimator which acts
as a control variate to reduce the overall variance of the policy
estimator.

\subsection{Hindsight Experience Replay} \label{sec:her}
Hindsight experience replay (HER) \cite{andrychowicz2017hindsight}
solves the sparse reward problem in RL by providing positive reward
even for failed states. In other words, it is motivated by the human
ability to learn useful things from failed attempts. Usually, HER is
applied to multi-goal environments where the agent is provided with
intermediate goals to achieve. Existing multi-goal environments such
as Mujoco's \texttt{FetchReach-v1} \cite{fetch} and PyBullet's
\texttt{Panda-Gym} \cite{gallouedec2021multi} do not provide image
observations which is necessitated by the scope of this study. Hence,
we propose three strategies to provide intermediate goals for the
problem environments considered in this paper. The first one is the
`success' strategy where the hindsight goal ($g^h$) is selected randomly from
a buffer containing previously encountered successful next states
($g^h=s': r=1$). The second strategy is called `final' state
strategy that uses the next state of the terminal step of the episode
($g^h=s' : d=1$) as the hindsight goal.  The third strategy is
called `future' state strategy where the hindsight goal is selected
randomly from future steps of the current buffer ($g_i^h=s_k' : i<k<n,
n=|\mathscr{B}|$). The agent receives an
intermediate reward of 1 when the L2-norm of distance between the current next
state and hindsight goal is less than a user-defined threshold.
Instead of calculating L2-distance directly between the images, it is
possible to compute the distance between the extracted features. This
is denoted by an additional letter `F' in the names used for labelling the plots. Mathematically, the hindsight reward for a given time step $i$ is given by:   
\begin{equation}
  r^h_i = \left\{\begin{array}{l} 1\quad \text{if}\; ||g^h_i - s_i'|| < h \\
               0\quad \text{otherwise}
  \end{array} \right.
  \label{eq:hind_reward}
\end{equation}
where $h$ is an user-defined threshold which kept at a value of 0.3 in
this paper. The implemented HER strategies is shown to provide
improvement over the IPG algorithm.

\subsection{Attention Architectures} \label{sec:attn_arch}
As mentioned earlier, attention  allows models to focus on
task-relevant aspects of the observations thereby providing robustness
against distractions and increased learning efficiency
\cite{vaswani2017attention} \cite{khan2021transformers}. The use of
attention mechanisms have been shown to improve the performance of
deep reinforcement learning algorithms
\cite{manchin2019reinforcement}. For instance, authors in
\cite{manchin2019reinforcement} have suggested several attention
mechanisms to improve RL performance in solving Arcade video games.
Similarly, authors in \cite{salter2019attention} use attention layers
along with privileged information (environment states) to improve RL
performance. Authors in \cite{iqbal2019actor} have used attention
along with an actor-critic model to solve a multi-agent RL problem. 

\begin{figure}[!t]
  \begin{tabular}{ll}
    \includegraphics[scale=0.2]{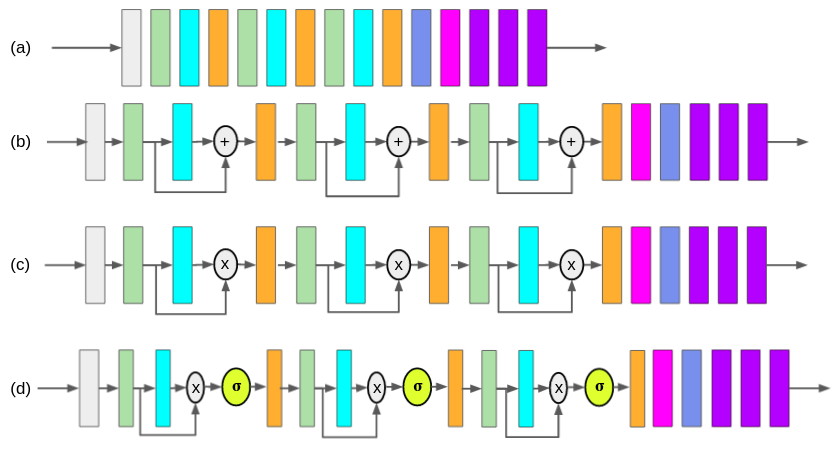} & \hspace{-0.2cm}
  \includegraphics[scale=0.2]{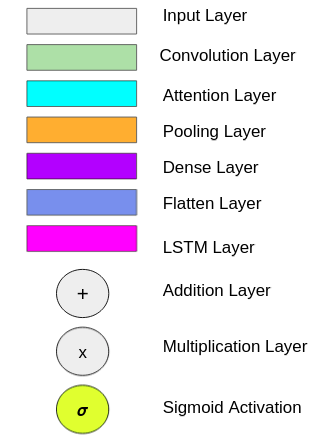}   
\end{tabular}
\caption{\footnotesize Feature Network Architectures with attention layers. (a) (\texttt{arch:0}) 
  attention layer (cyan) is placed after each  convolutional layer
  (green), (b) (\texttt{arch:1}) Output of the attention layer is added to its input
  before being passed to the next layer: $y=x+Attention(x)$. (c) (\texttt{arch:2}) Output
of the attention layer is multiplied to its input: $y=x*Attention(x)$
(d) (\texttt{arch:3}) the output of architecture (c) is passed through a sigmoid
activation function: $y = \sigma(x * Attention(x))$.}
  \label{fig:attn_arch}
\end{figure}

In this paper, we apply attention to two vision-based robotics
problems. Two types of attention mechanisms, namely, Bahdanau's
additive type \cite{bahdanau2014neural} and Luong's dot product type
\cite{luong2015effective} are used in four different configuration as
shown in Figure \ref{fig:attn_arch}. The attention layers are used within a 
feature network which is shared by both actor and critic models. An
optional LSTM layer is also used along with the attention layer to
capture the temporal relationships between the frames. The effect of
attention layers on learning performance of RL algorithms is discussed
in the next section.

\section{Experiments}\label{sec:expt}
The details of experiments carried out and the analysis of results are
presented in this section as described below.

\subsection{Software and Hardware Configurations}
The algorithms were implemented using Python 3.7 and Tensorflow 2.4 on
Ubuntu 20.04 GNU/Linux operating system. The source codes are made
available publicly on github \cite{swgrlcode} for the convenience of
readers and to ensure reproducibility of results. The plots were drawn
using wandb machine learning platform \cite{wandb}. The pre-built
simulation environments available with Pybullet \cite{pybullet} were
used for generating the results presented in this paper. The problem
environments that are included this study are \texttt{KukaDiverseObjectEnv} and
\texttt{RaceCarZEDGymEnv}. In these environments, the state
observation is available in the form of RGB images and the action
space is continuous, thereby making them suitable for this study which
concerns itself with the hand-eye coordination or vision-based
robotics problem. Each run of the program takes about 4-7 hours on a
GeForce RTX 2060 GPU machine with about 6 GB of dedicated video RAM.
Various hyper-parameters used for different algorithms is shown in
Table \ref{tab:hparam}. Each season corresponds to 1024 time-steps
which is approximately about 140 episodes with 7-8 steps per episode
in case of Kuka environment. This is also the length of trajectory
segment (denoted by $|\mathscr{D}|$) which is used for on-policy
training algorithms such as PPO. Each of the algorithm is executed at
least for 5 runs with random seed initialization to generate the
statistical plots. The attention and LSTM layers are implemented using
Keras APIs. 

\begin{table}[htbp]
  \centering
  {\tiny
  \begin{tabular}{|l|l|} \hline
    Parameters &  Value     \\ \hline
    Replay Buffer Size $|\mathscr{B}|$ & 20,000   \\ \hline
    Batch size & 128  \\ \hline
    Training epochs & 20  \\ \hline
    Trajectory Length (On-policy buffer) $|\mathscr{D}|$ & 1024  \\ \hline
    Discount factor $\gamma$ & 0.995  \\ \hline
    Learning rate $\eta$ & 0.002  \\ \hline
    Input stack size & 7  \\ \hline
    Polyak averaging factor $\tau$ & 0.995  \\ \hline
    Clip Factor in PPO $\epsilon$ & 0.2  \\ \hline
    Discount factor in GAE $\lambda$ & 0.7 \\ \hline
    Entropy Coefficient in SAC $\alpha$ & 0.2 \\ \hline
  \end{tabular}}
  \caption{\footnotesize Hyper-parameters used for different RL algorithms. }
  \label{tab:hparam}
\end{table}
\subsection{Benchmarking of RL Algorithms}
\begin{figure}[htbp]
  \centering
  \begin{tabular}{cc}
    \includegraphics[scale=0.1]{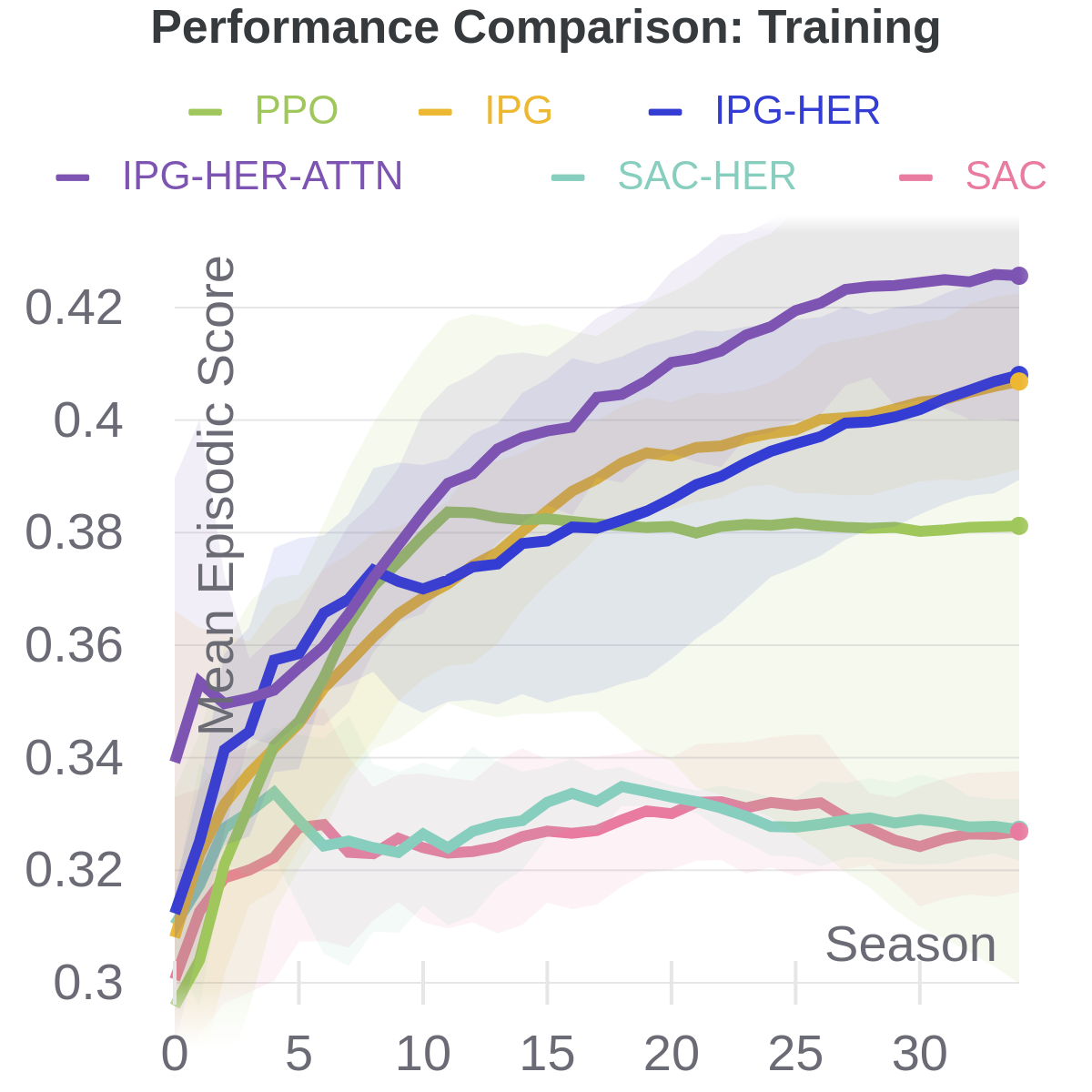} & 
    \includegraphics[scale=0.1]{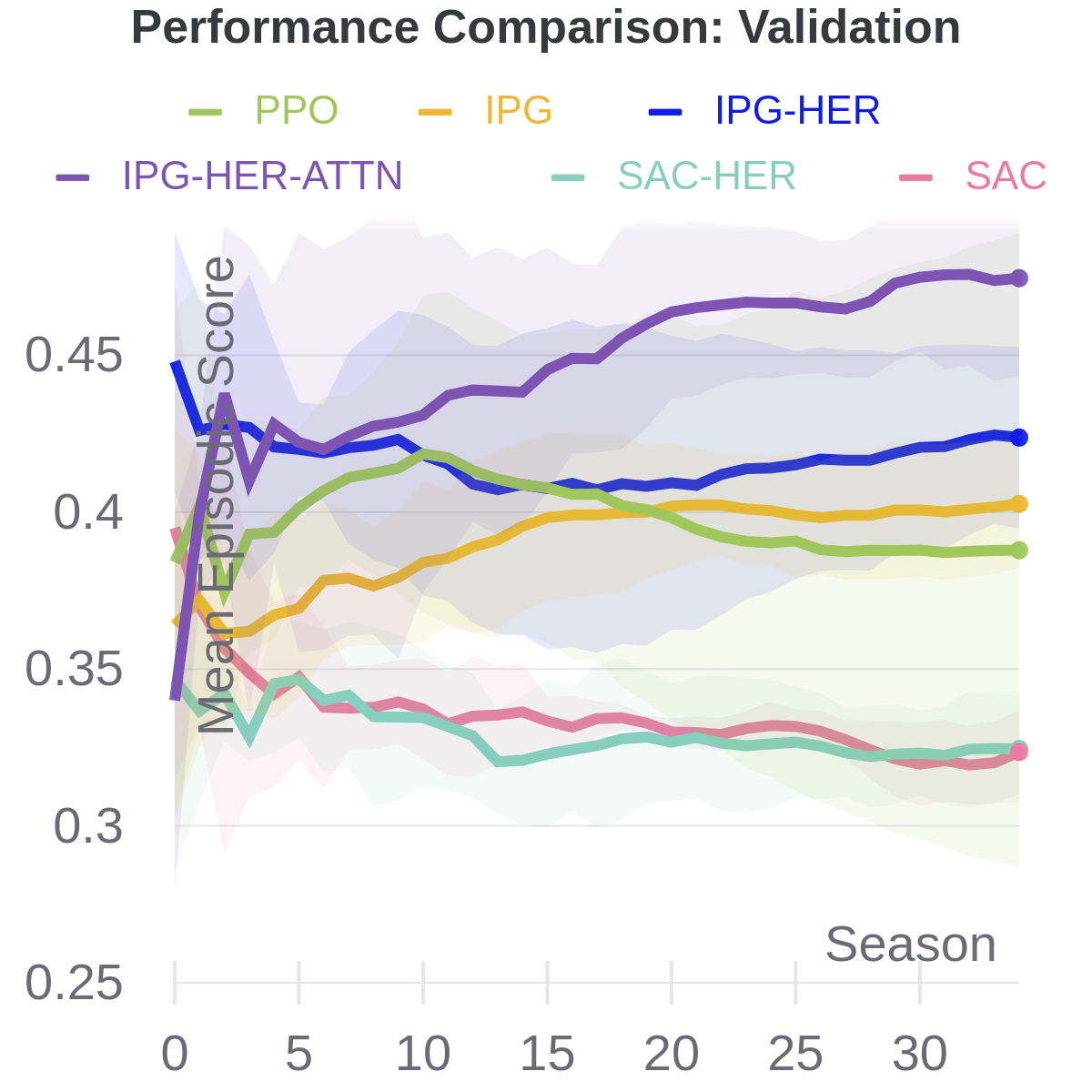} \\
    {\scriptsize (a) Training} &  {\scriptsize (b) Validation}
  \end{tabular}
  \caption{\footnotesize Performance Comparison of various RL methods for
    \texttt{KukaDiverseObject} Environment. The mean episodic score
    average over seasons is shown on the y-axis. Each season corresponds
    to 1024 time-steps or about 140 episodes. Validation involves
  computing rewards for 50 episodes with deterministic action policy.
The off-policy and on-policy training steps are mapped to seasons for
comparison.}
  \label{fig:kuka_perf_comp}
\end{figure}

\begin{figure}[htbp] \centering
  \includegraphics[scale=0.35]{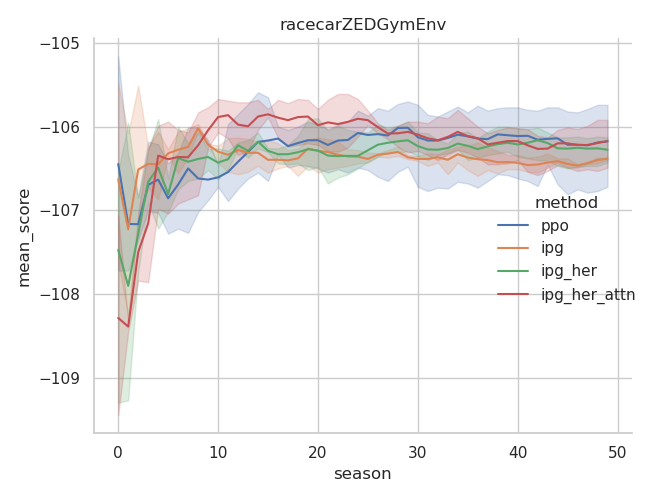}
  \caption{\footnotesize Performance Comparison of RL algorithms for
    \texttt{RaceCarZEDGymEnv}. The maximum number of steps in an
    episode is limited to 20. The input images are resized to
    (40,40,3) before its use in the training. The y-axis shows the
  mean episodic score against the seasons on x-axis. }
  \label{fig:rc_perf_comp}
\end{figure}

The performance of four state-of-the-art RL algorithms namely, Soft
Actor Critic (SAC) \cite{haarnoja2018soft}, Proximal Policy
Optimization (PPO) \cite{schulman2017proximal}, Interpolated Policy
Gradient (IPG) \cite{gu2017interpolated} and Hindsight Experience
Replay (HER) \cite{andrychowicz2017hindsight} is compared for two
OpenAI/Gym simulation problem environments, namely,
\texttt{KukaDiverseObjectEnv} and \texttt{RaceCarZEDGymEnv}
respectively. The outcome is shown in Figures \ref{fig:kuka_perf_comp}
and \ref{fig:rc_perf_comp} respectively. One can observe that IPG
provides superior performance compared to SAC and PPO. The HER variant
of IPG, called, IPG+HER provides better performance compared to IPG.
It is also shown that the performance can be improved further by using
attention layers. The results are consistent for both the example
problems.

\subsection{Effect of Attention on learning performance}
\begin{figure}[htbp]
  \centering
  \begin{tabular}{cc}
    \includegraphics[scale=0.1]{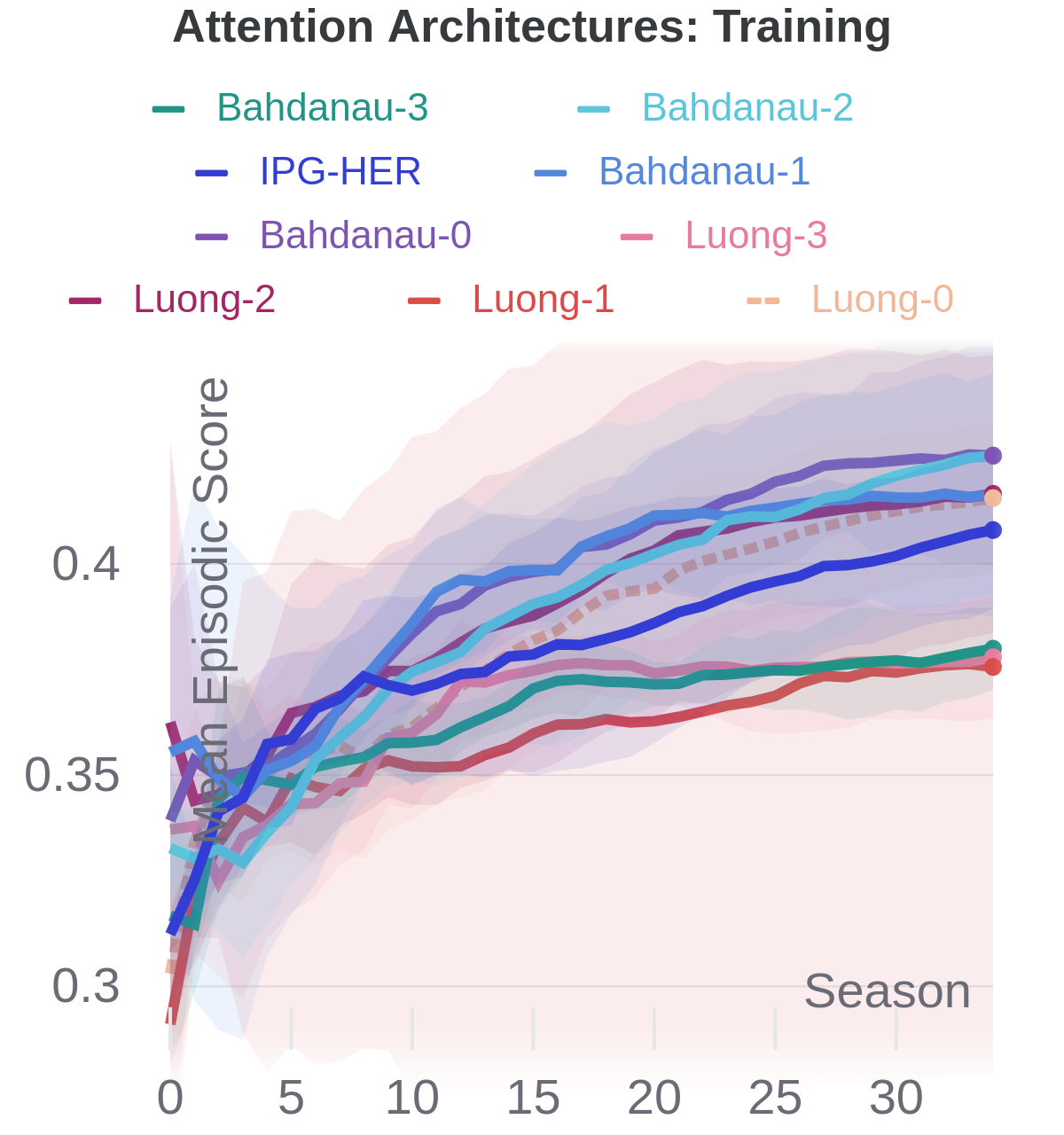} &
    \includegraphics[scale=0.1]{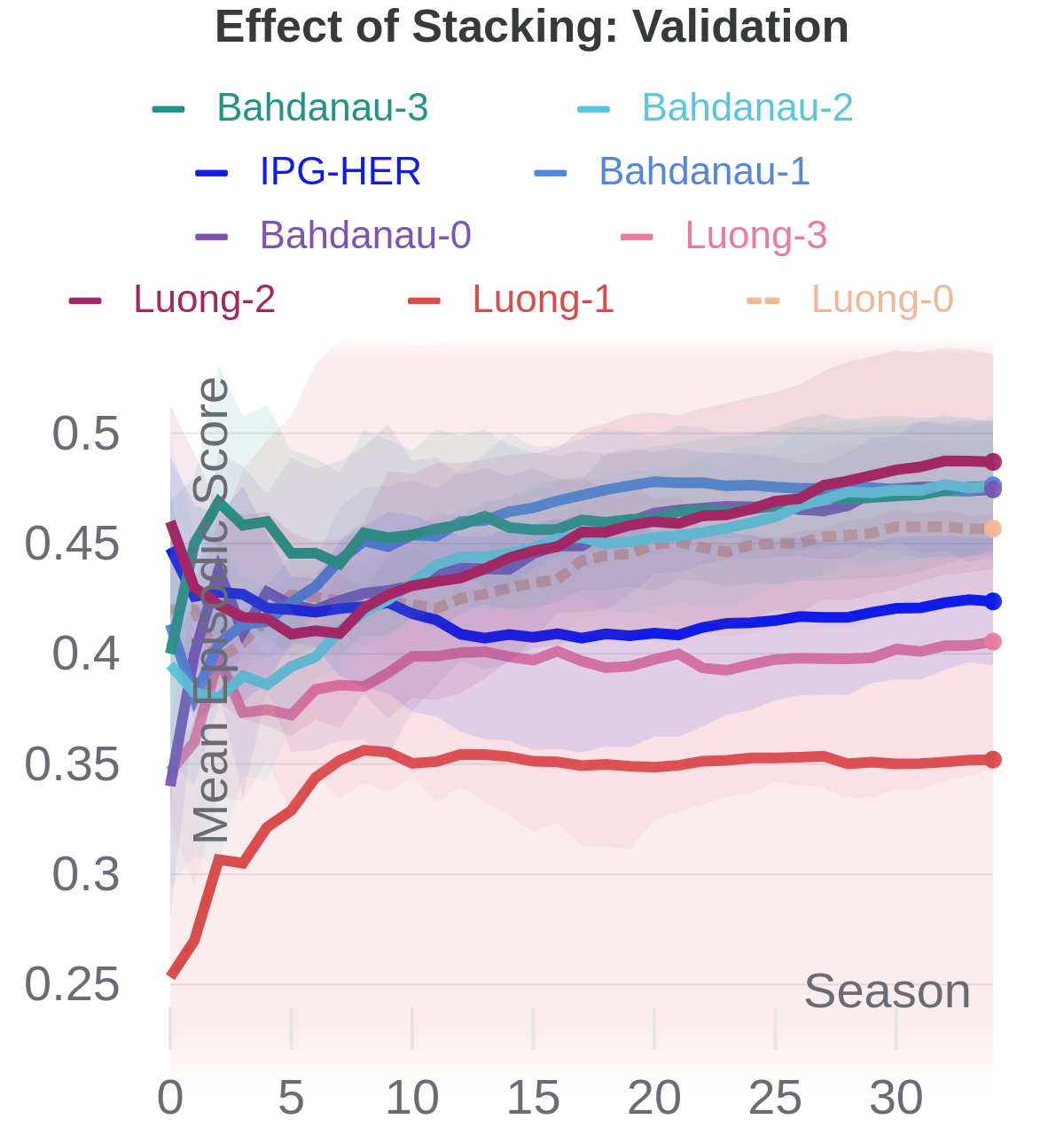}  \\
    {\small (a) Train} & {\small (b) Validation}
  \end{tabular}
  \caption{\footnotesize Effect of attention mechanisms and architectures on
    the learning performance of IPG-HER algorithm. The IPG-HER
    algorithm without attention acts as a baseline for comparison
    (shows in dark blue). Two attention types, namely, Bahdanau and
    Luong, are considered. Four architectures are shown in Figure
    \ref{fig:attn_arch}. Other
    hyper-parameters are: {\texttt{her\_strategy:`future',
    stack\_size=1}}.} 
    \label{fig:attn_comp}
\end{figure}

Figures \ref{fig:kuka_perf_comp} and \ref{fig:rc_perf_comp} show
that it is possible to obtain better learning performance using
attention layers. The effect of various attention architectures
discussed in Section \ref{sec:attn_arch} is shown in Figure
\ref{fig:attn_comp}. It is seen that the architectures
\texttt{`Luong:2', `Bahdanau:0', 'Bahdanau:2'} provide considerable
improvement over the base algorithm (IPG-HER) in case of \texttt{KukaDiverseObjEnv}.

\begin{figure}[htbp]
  \centering
  \begin{tabular}{cccccc}
    \includegraphics[scale=0.6]{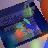} &
    \includegraphics[scale=0.6]{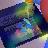} &
    \includegraphics[scale=0.6]{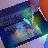} &
    \includegraphics[scale=0.6]{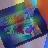} &
    \includegraphics[scale=0.6]{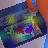} &
    \includegraphics[scale=0.6]{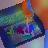} \\
    \includegraphics[scale=0.6]{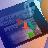} &
    \includegraphics[scale=0.6]{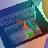} &
    \includegraphics[scale=0.6]{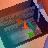} &
    \includegraphics[scale=0.6]{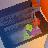} &
    \includegraphics[scale=0.6]{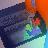} &
    \includegraphics[scale=0.6]{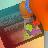}
  \end{tabular} \caption{\footnotesize Gradient Content Activation Map attention
    Layers. The top row shows the output of the second attention layer
  and the bottom row shows the output of the outermost attention
layer.} \label{fig:gradcam} \end{figure}

The output of different attention layers is visualized using grad-CAM
\cite{selvaraju2017grad} is shown in Figure \ref{fig:gradcam}. The
highlighted regions in the image are the areas which were given more
importance for computing actions for that particular state. 
                                                                   
\subsection{Effect of HER strategies} \label{sec:abln}
\begin{figure}[htbp]
  \centering
  \begin{tabular}{cc}
    \includegraphics[scale=0.07]{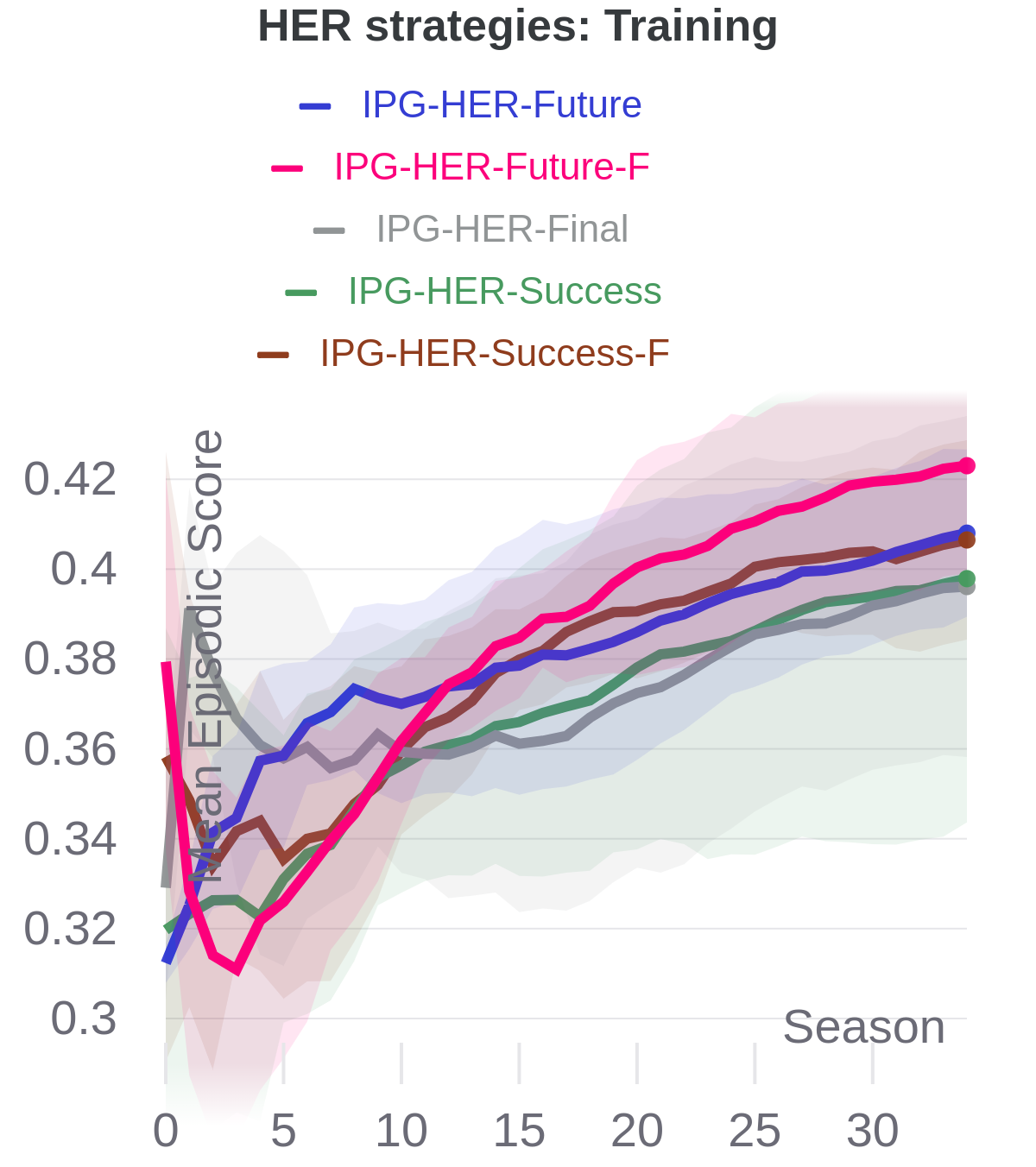} & 
    \includegraphics[scale=0.07]{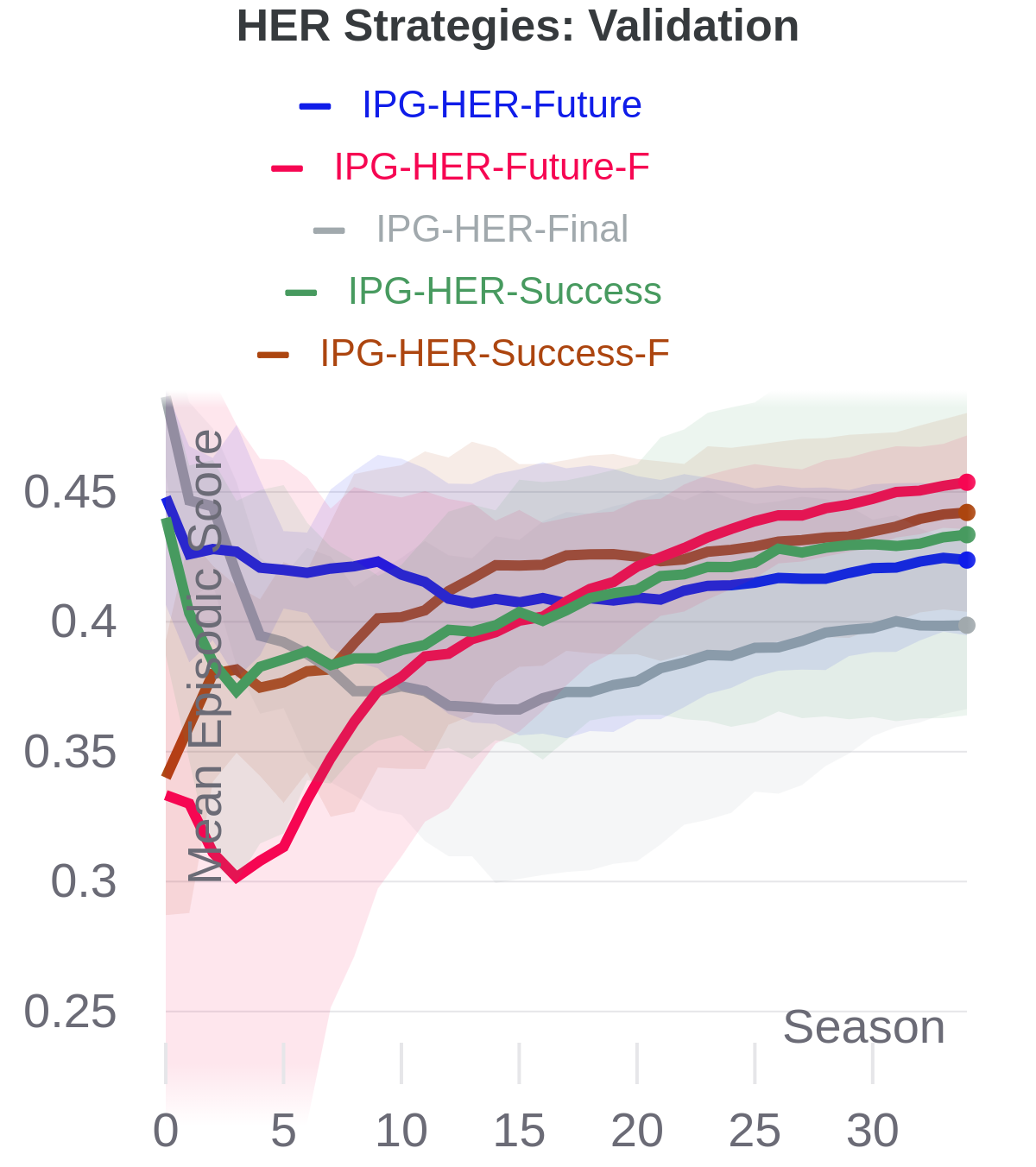} \\
    {\small (a) Train }  &  {\small (b) Validation } 
  \end{tabular}
  \caption{\footnotesize Effect of goal selection in HER on learning performance.
    The letter `F' indicates that the L2-distance is computed for the
    extracted features of the state and the goal images instead of using
  the raw images.} 
  \label{fig:her_goal}
\end{figure}

The effect of various strategies for selecting intermediate goals on
RL performance is shown in Figure \ref{fig:her_goal}. It is seen that
the `future' strategy provides superior performance compared to other
strategies. Further improvement is obtained when the extracted
features are used for computing the intermediate rewards for the
agent. This is denoted by the letter `F' in the algorithm names
mentioned in the legend.  

\subsection{Effect of using stacked Frames} \label{sec:stack}
\begin{figure}[htbp]
  \centering
  \begin{tabular}{cc}
    \includegraphics[scale=0.07]{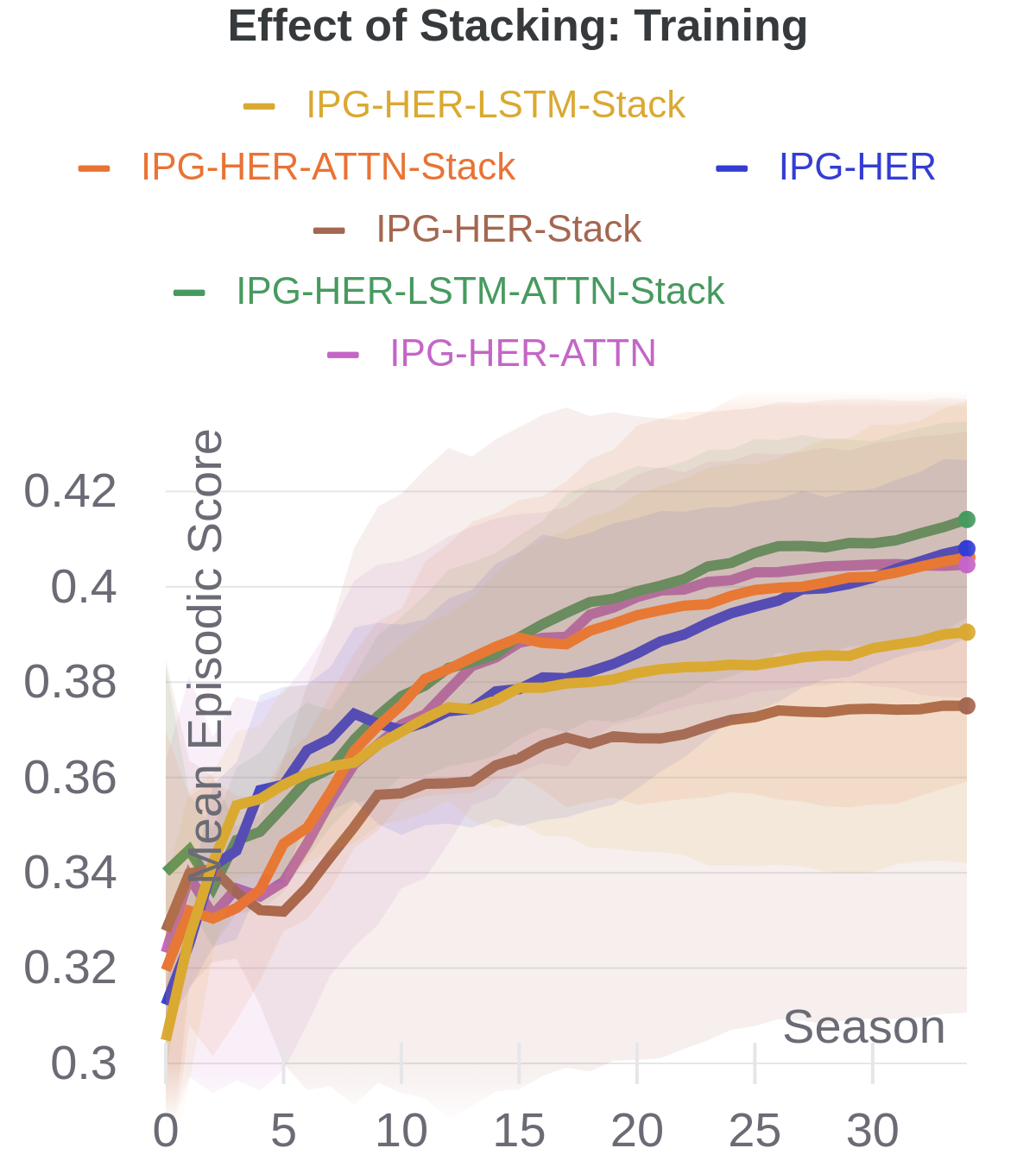} & 
    \includegraphics[scale=0.07]{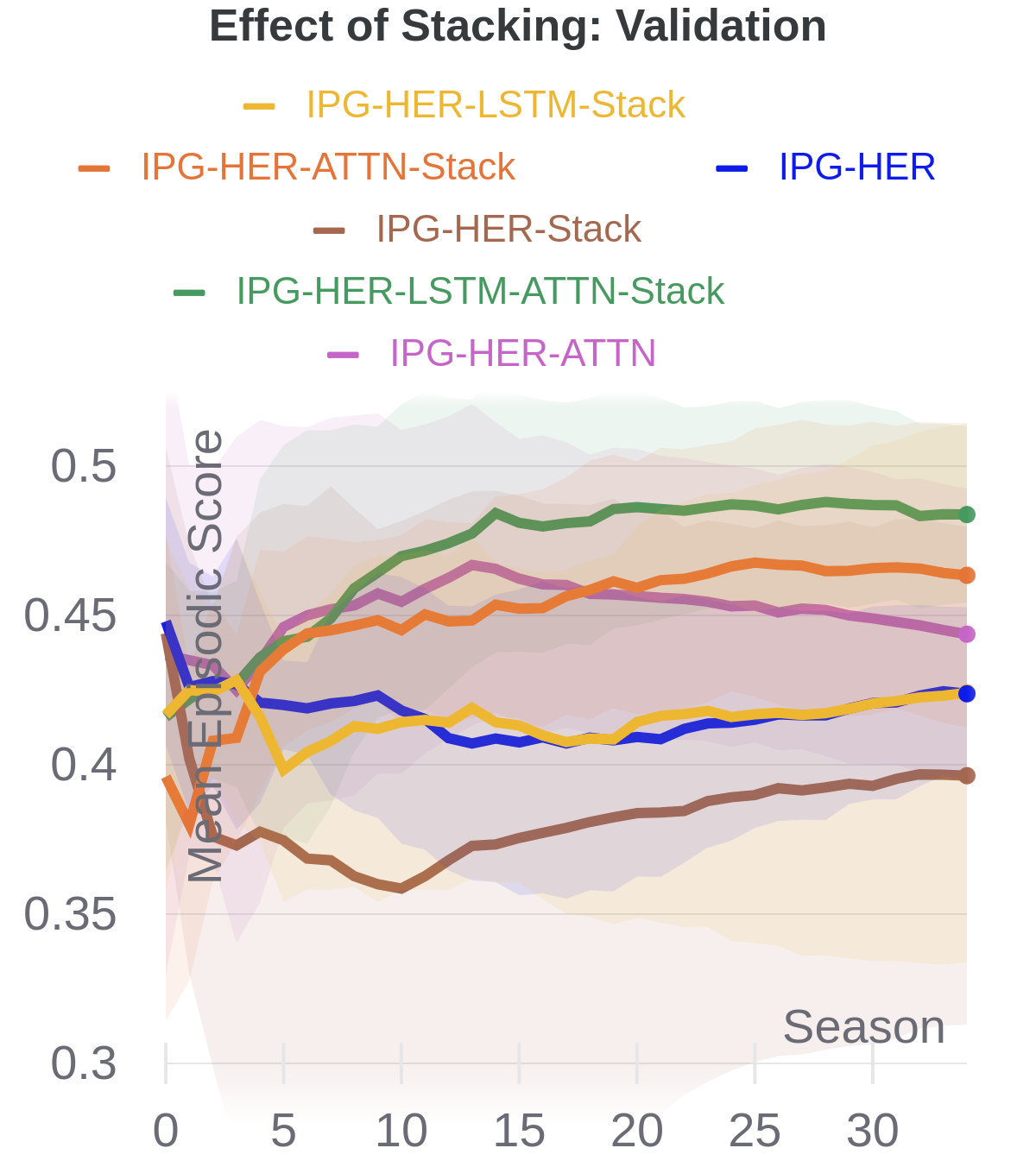} \\
    {\small (a) Training} & {\small (b) Validation}
  \end{tabular}
  \caption{\footnotesize Effect of Stacking on performance of IPG-HER
    algorithm for \texttt{KukaDiverseObjectEnv}. It is observed
    that attention is more effective in improving RL performance with
    stacked input frames. LSTM is also found to improve the learning
    performance with stacking. The best performance is obtained by
    combining attention with LSTM and stacking. Hyper-parameters used
    are: \texttt{attn\_arch= Luong:0, stack\_size=7,
    her\_strategy=`future'}.} 
  \label{fig:stack_comp} 
\end{figure}

The temporal relationships present in the state input can be taken
into consideration by stacking the frames together and is known to
overcome the partial observability problem in a Markov decision
process (MDP) \cite{kapturowski2018recurrent} \cite{manchin2019reinforcement}. 
The effect of using stacked frames on the learning performance is shown in
Figure \ref{fig:stack_comp}. It is observed that stacking itself does
not improve the learning performance (blue vs brown curves). However
it improves the learning performance when used along with attention
and/or LSTM layers. The best performance is obtained when LSTM,
Attention and Stacking are used together. It can be concluded that
incorporating spatial and temporal attention can improve the
performance of RL algorithms.  


\section{Conclusion}\label{sec:conc}
This paper presents a benchmarking study of some of the
state-of-the-art reinforcement learning algorithms for two simulated
vision-based robotics problems namely, \texttt{KukaDiverseObjectEnv}
and \texttt{RaceCarZedGymEnv}. In both of these two environments, the
observation is available in the form of RGB images and the action
space is continuous making them suitable for this study. The
algorithms considered for benchmarking include Soft Actor Critic
(SAC), Proximal Policy Optimization (PPO), Interpolated Policy
Gradients (IPG) and Hindsight Experience Replay (HER). A number of
strategies are suggested to select intermediate goals required for
implementing HER algorithm on these single-goal problem environments.
In addition, it is shown that the learning performance of RL
algorithms could be enhanced further by applying spatial and temporal
attention. To the best of our knowledge such a benchmarking study for
these two environments are not available in the literature making it a
novel contribution in the field. The future direction would be to
implement these algorithms to multi-goal environments with image
inputs, which are now becoming available recently. Another direction
would be to extend this benchmarking to other robotic problems. 


\bibliographystyle{ieee} 
\bibliography{ref} 

\end{document}